\documentclass{article}

\usepackage[preprint]{neurips_2025}

\usepackage[utf8]{inputenc} 
\usepackage[T1]{fontenc}    
\usepackage{hyperref}       
\usepackage{url}            
\usepackage{booktabs}       
\usepackage{amsfonts}       
\usepackage{nicefrac}       
\usepackage{microtype}      
\usepackage{xcolor}         
\usepackage{graphicx}
\usepackage{wrapfig}
\usepackage{amsmath}
\usepackage{adjustbox}
\usepackage{multirow}
\usepackage{lipsum}
\usepackage{amsfonts}       
\usepackage{xspace}         
\usepackage{wrapfig,lipsum}
\usepackage{colortbl}
\usepackage[capitalize]{cleveref}  
\crefname{section}{Sec.}{Secs.}
\Crefname{section}{Section}{Sections}
\crefname{table}{Tab.}{Tabs.}
\Crefname{table}{Table}{Tables}
\crefname{figure}{Fig.}{Figs.}
\Crefname{figure}{Figure}{Figures}
\crefname{equation}{Eq.}{Eqs.}
\Crefname{equation}{Equation}{Equations}
\newcommand{\method}{\texttt{ROSE}\xspace}

\title{ROSE: Remove Objects with Side Effects in Videos}

\author{%
  Chenxuan Miao$^1$, Yutong Feng$^{2}$\thanks{Project Leader}, Jianshu Zeng$^3$, Zixiang Gao$^3$, Hantang Liu$^2$, \\
  \vspace{3pt}
  \textbf{Yunfeng Yan$^1$, Donglian Qi$^1$, Xi Chen$^4$, Bin Wang$^2$, Hengshuang Zhao$^{4}$\thanks{Corresponding Author}}\\
  \vspace{3pt}
  $^1$Zhejiang University, $^2$KunByte AI, $^3$Peking University, $^4$The University of Hong Kong$$\\
  \texttt{\{weiyuchoumou526, fengyutong.fyt, zengjianshu.AI, gzx2401210062
  \}@gmail.com}
  \\ \texttt{\{liuhantang77, chauncey0620, binwang393\}@gmail.com} \\
  \texttt{\{yyff, qidl\}@zju.edu.cn} \quad \texttt{hszhao@cs.hku.hk}
}

\begin{document}

\maketitle

\begin{abstract}
Video object removal has achieved advanced performance due to the recent success of video generative models.
However, when addressing the side effects of objects, \textit{e.g.,} their shadows and reflections, existing works struggle to eliminate these effects for the scarcity of paired video data as supervision.
This paper presents \method, termed \textbf{R}emove \textbf{O}bjects with \textbf{S}ide \textbf{E}ffects, a framework that systematically studies the object's effects on environment, which can be categorized into five common cases: shadows, reflections, light, translucency and mirror.
Given the challenges of curating paired videos exhibiting the aforementioned effects, we leverage a 3D rendering engine for synthetic data generation.
We carefully construct a fully-automatic pipeline for data preparation, which simulates a large-scale paired dataset with diverse scenes, objects, shooting angles, and camera trajectories.
\method is implemented as an video inpainting model built on diffusion transformer.
To localize all object-correlated areas, the entire video is fed into the model for reference-based erasing.
Moreover, additional supervision is introduced to explicitly predict the areas affected by side effects, which can be revealed through the differential mask between the paired videos.
To fully investigate the model performance on various side effect removal, we presents a new benchmark, dubbed ROSE-Bench, incorporating both common scenarios and the five special side effects for comprehensive evaluation.
Experimental results demonstrate that \method achieves superior performance compared to existing video object erasing models and generalizes well to real-world video scenarios. The project page is \url{https://rose2025-inpaint.github.io/}.

\end{abstract}

\begin{figure}[htbp]
    \centering
    \includegraphics[width=0.9\linewidth]{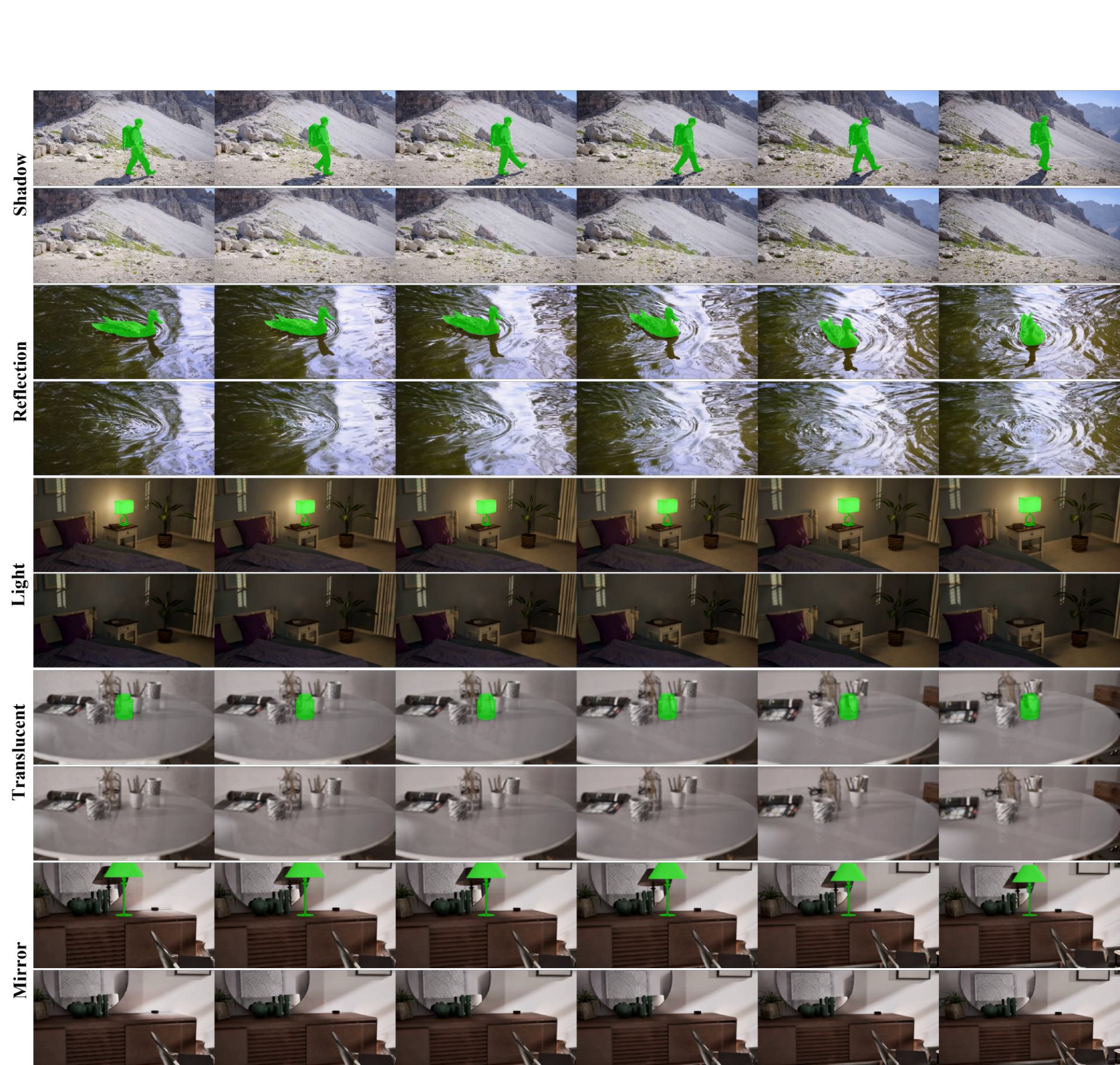}
    \vspace{-5pt}
    \caption{\textbf{Video object removal results generated by \method} (zoom in for better view). Every two lines are an example where the above is input video with mask and the bottom is inference result. We sequentially show cases of various side effects studied in this paper.}
    \label{ours_demo}
    \vspace{-18pt}
\end{figure}

\section{Introduction}
    
Removing objects in visual contents represents a valuable technique with widespread applications in both daily and industrial scenarios.
This task targets to re-fill the masked region of objects via reasonable and consistent content, regarding the context in surrounding environment.
Prior works~\cite{ju2024brushnet,wei2025omnieraser,jiang2025smarteraser,bian2025videopainter} towards either image or video object removal have explored to leverage flow-based pixel propagation to restore the masked region with neighboring information~\cite{zhou2023propainter}, or adopt the inpainting paradigm to directly generate the masked content~\cite{LAMA,li2025diffueraser}. 
Powered by the significant capability of large-scale models~\cite{sdxl,sd3,flux2024,li2025diffueraser} on generalized visual creation, the inpainting-based methods exhibit satisfying erasing performance in diverse scenarios of image and video.


Despite the advanced performance, however, existing works are still restricted due to the lack of paired training samples that follows real-world physical rules.
The paired samples represents data with and without the object, where the object's influence on the environment is correspondingly changed, such as its shadow on the ground. 
Most works leverage the segmentation dataset, \textit{e.g.,} DAVIS~\cite{davis} and YouTube-VOS~\cite{xu2018youtube}, to construct artificial pairs, either directly pasting an object from another sample, or masking the object with zero value.
While simple and scalable, these strategies fail to reflect the side effects of object, \textit{e.g.,} shadows, reflections and lighting changes.
Therefore, models supervised by the artificial pairs typically generate unnatural outputs with side effects left in environment.
To tackle this, OmniEraser~\cite{wei2025omnieraser} manages to filter out such image pairs from the sequential frames in videos with static camera motion.
However, when confronting with video object removal, it is impractical to leverage higher dimensional data to construct the paired dataset.


To address these problems, we propose to prepare the paired video samples via 3D rendering.
Recent advancements on the rendering engines~\cite{epic2024unreal} make it practical to generate high-qualified and strictly-aligned synthetic video pairs.
We design a fully-automatic data preparation pipeline to create a large-scale video set for object removal.
More concretely, we collect a batch of base environments and split them into multiple scenes containing various objects. 
Following that, the pipeline automatically generates cameras focusing on the objects to be removed, and apply random camera trajectories. 
The rendering engine enables us to activate or disable the object, and also precisely render the object masks.
Thus, we could obtain a list of triples consisting of the original video, edited video with object removed, and the corresponding mask video, which contain perfectly simultaneous temporal contents.
Furthermore, we systematically study the various types of side effects in videos, including light source, mirror, reflection, shadow, and translucency.
Equipped with the data preparation pipeline, we efficiently construct a comprehensive dataset including all the above side effects on diverse scenes.


To fully utilize the synthetic data, we present \method, an efficient framework based on video inpainting to remove object in videos with their side effects.
To help distinguish the object-interacted region in environment, we directly feed the whole video into the model, in contrast to previous works that fills the object area with zero mask.
The complete video serves as a powerful reference guidance on model, to localize the side effects concerning the intrinsic attributes of the object.
We also apply random augmentation strategies on the mask to cope with various input in inference.
Furthermore, we introduce an additional supervision to explicitly predict the difference mask between edited and original videos. 
We implement this by injecting a mask predictor based on the hidden representations of the inpainting model.
The aforementioned architectures of \method are observed to enhance the model's capability to attend and erase the side effects in videos.


To facilitate a comprehensive evaluation on the object removal results with side effects, we construct a new benchmark, named ROSE-Bench, consisting of both realistic and synthetic video data.
Through extensive experiments, we demonstrate that \method achieves state-of-the-art performance on video object removal, and effectively adapts to real-world scenarios.
    

\section{Related Work}
\label{gen_inst}

\paragraph{Diffusion Transformers for Video Generation. }

Recent diffusion models~\cite{ho2020ddpm,rombach2022high,sohl2015deep,song2021scorebased} have shown strong performance in text-to-video generation. By integrating transformers~\cite{vaswani2017attention}, diffusion transformers (DiTs)\cite{peebles2023dit} improve video quality and temporal consistency. State-of-the-art methods leverage large-scale video-text datasets\cite{bain2021frozen,xu2023video} and hybrid architectures for efficiency and fidelity. Recent DiT-based latent diffusion models, such as Wan2.1~\cite{wan2025video} and MAGI-1~\cite{magi2025video}, excel in long video generation: Wan2.1 uses causal 3D VAEs with $1{:}256$ compression and flow matching for real-time synthesis, while MAGI-1 employs an autoregressive DiT for chunk-wise generation with strict causality. These advances underscore DiTs' strength in balancing quality, efficiency, and control.

\vspace{-6pt}

\paragraph{Video Inpainting. }
Early video inpainting methods primarily used 3D CNNs~\cite{freeformvideoinpainting,videoinpaintingbyjointlylearning,proposalbasedvideocompletion} to model spatial-temporal features, but their limited receptive fields hindered long-range propagation. Subsequently, optical flow~\cite{kim2019deepvideoinpainting,li2020shortlongterm,zou2021progressive} and homography~\cite{lee2019copypaste,cai2022devit} were introduced to guide pixel propagation. To improve efficiency and accuracy, Zhou et al.\cite{zhou2023propainter} proposed ProPainter, combining optical flow and attention mechanisms. Recently, with the rise of diffusion models\cite{ho2020ddpm,rombach2022high}, diffusion-based video inpainting has emerged~\cite{shi2024bivdiff,lee2024video,wu2024towards,zi2024cococo}. Li et al.\cite{li2025diffueraser} proposed DiffuEraser, extending the image inpainting model BrushNet\cite{ju2024brushnet} to videos via a two-stage training scheme.

\section{Dataset Construction}
\label{dataset_construct}

\subsection{Paired Erasing Videos Preparation using 3D data}

Acquiring paired data samples that depict scenes with and without objects and their side effects represents a significant challenge in object removal task.
Though recent work explores generating such image pairs from videos with static camera motion \cite{wei2025omnieraser}, it is impossible to obtain video pairs 
 in a higher dimension using this technique.
To tackle this problem, we propose to utilize the adequate 3D data together with advanced game engine, \textit{i.e.,} the Unreal Engine~\cite{epic2024unreal}, to synthesize the paired video data. 
As illustrated in \cref{Data_preparation_pipeline}, we present an automatic data preparation pipeline as follows:

\begin{figure}[t]
    \centering
    \includegraphics[width=0.9\textwidth]{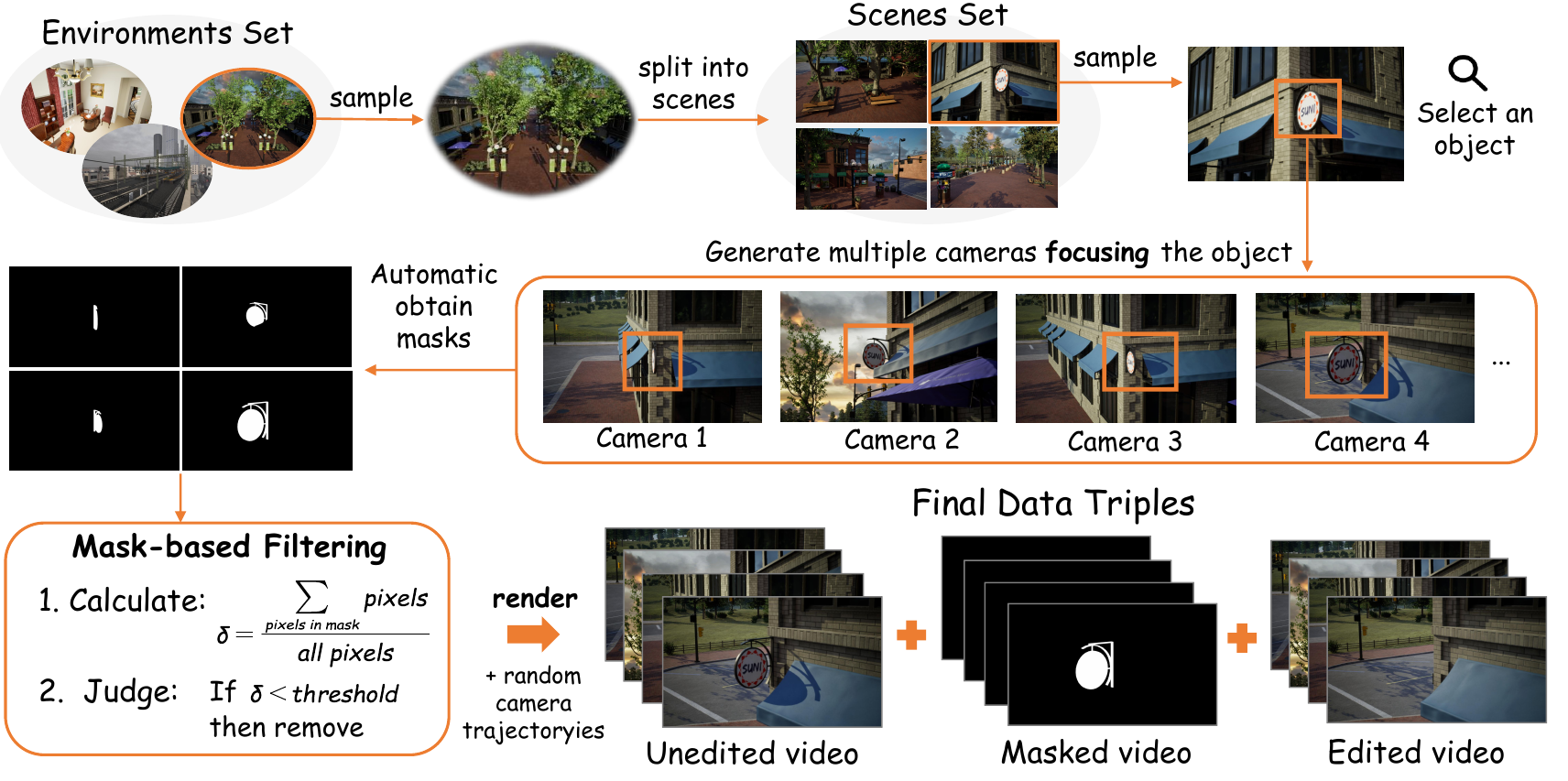}
    \vspace{-5pt}
    \caption{\textbf{Paired video preparation pipeline using 3D data}, which can be divided into: scene and object sampling, multi-view generation with masks, valid view filtering and video data rendering.}
    \label{Data_preparation_pipeline}
    \vspace{-15pt}
\end{figure}

\textbf{Scene and Object Sampling.}
We begin by collecting large-scale virtual environments from public 3D asset platforms such as Fab~\cite{fab}. Each environment is sufficiently complex and diverse, covering a wide range of indoor and outdoor scenes, including urban settings, natural landscapes, and artificial constructions. We manually subdivide these base environments into smaller scenes, each containing one or more candidate objects for removal. In total, we collect 28 high-quality environments and split them into 450 unique scenes. The selected scenes include a wide variety of object types—both static and dynamic—including vehicles, animals, plants, and more. This ensures a diverse training corpus that enhances the generalization ability of the inpainting model.
    
\textbf{Multi-view Generation with Object Masks.}
Given a sampled object in a scene, we randomly assign multiple camera views with varying angles and distances within predefined ranges.
A key advantage of using a 3D engine is the ability to generate accurate object masks via programmable post-process shaders, avoiding reliance on segmentation models~\cite{sam}.
For each object, we apply a custom shader that renders the object in white and masks the rest in black, producing precise binary masks.
Per-frame mask videos are automatically generated through scripting.

\textbf{Valid View Filtering.}
To ensure the quality of videos and avoid object-occlusion cases, we further filter out views by calculating the ratio of foreground pixels in the mask. 
Ratios lower than a threshold suggest videos with insufficient mask coverage, \textit{e.g.,} due to occlusion or mislabeling.
Such videos are discarded to avoid introducing noisy supervision into the training set.
    
\textbf{Video Pair Rendering.}
After filtering, we render both the unedited (original) and edited (object-removed) video sequences by toggling the visibility of the selected objects in the engine. 
The camera moving is sampled from a pre-defined set with random disturbing, \textit{e.g.,} zooming in and out.
All video pairs are rendered at a resolution of $1920\times 1080$ and a frame length of 90 frames (6 seconds). 
Since the camera trajectories and object placement are determined via scripted generation, the original video, the corresponding mask video, and the edited video remain spatially and temporally aligned on a per-frame basis. 
Such an alignment is critical for enabling pixel-wise supervised learning.

\subsection{Categorize Side Effect in Videos}

    To improve the generalization ability of the model and its robustness under various complex real-world conditions, we deliberately construct the dataset composed of six distinct categories. These categories are carefully designed to simulate typical yet challenging side effects that commonly occur in practical scenarios, such as object-light interactions, mirror reflections, and translucent materials. By explicitly injecting such variations into the training process, we aim to equip the model with the capacity to understand and handle diverse object-environment relationships beyond trivial inpainting cases. We summarize the definition of side effects on the environment as follows:

                \begin{figure}[bp]
        \centering
        \vspace{-10pt}
        \includegraphics[width=0.95\linewidth]{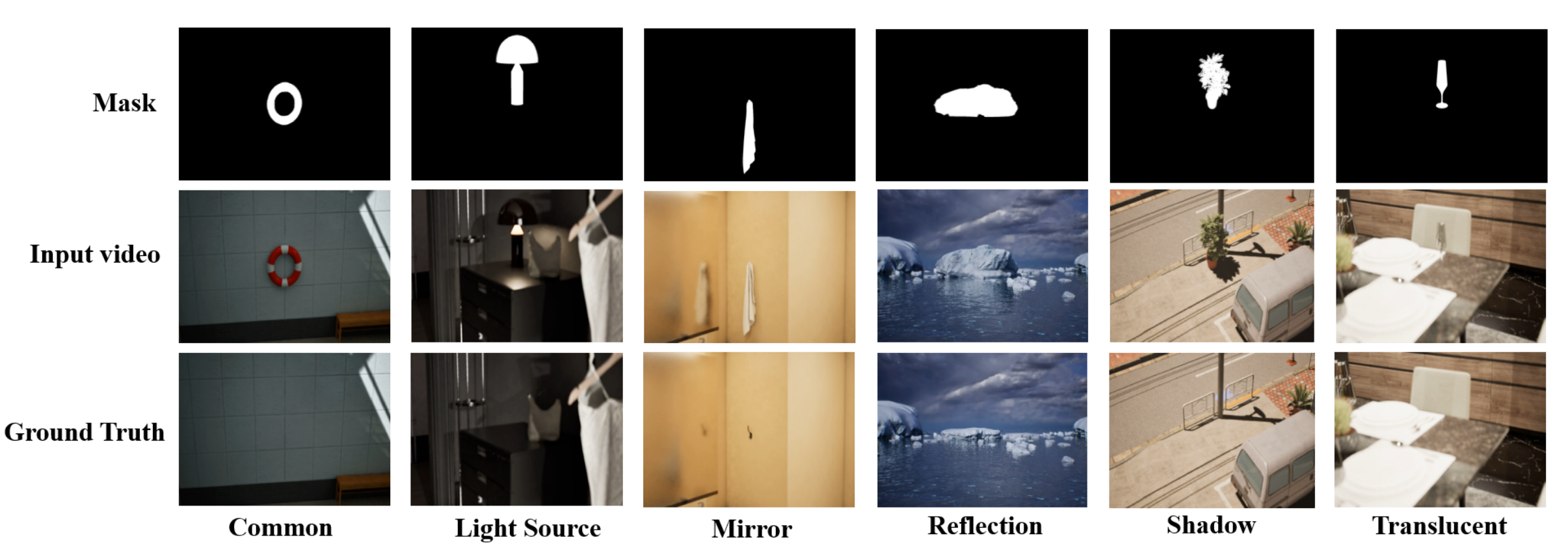}
        \vspace{-10pt}
        \caption{Illustration of the various \textbf{side-effect categories} studied in the dataset of \method.}
        \label{dataset_demo}
        \vspace{-10pt}
    \end{figure}
    
        \textbf{Common}: 
        Objects with minimal interaction with surrounding context, representing typical inpainting cases. Their removal causes little disruption to spatial layout or visual semantics.

        \textbf{Light Source:} 
        This category includes objects that function as light emitters. Their removal changes the global illumination, affecting shadows, reflections, and overall scene appearance.

        \textbf{Mirror:} 
        Objects reflected in mirrors require spatial reasoning and semantic understanding to inpaint both the object and its mirrored counterpart, ensuring visual consistency.

        \textbf{Reflection:}
        Compared to the Mirror category, it emphasizes reflective surfaces like water, requiring the model to infer and complete indirect visual cues from reflections.
        
        \textbf{Shadow:} Shadows linked to objects require joint removal, making inpainting sensitive to lighting and spatial structure to ensure coherence across both object and shadow regions.
        
        \textbf{Translucent:} Semi-transparent objects expose the background with blending or refraction. Inpainting must recover both visible cues and hidden structures for realistic restoration.

\section{Method}
\label{headings}
\subsection{Overview}

\begin{figure}[t]
        \centering
        \includegraphics[width=\linewidth, scale=1.2]{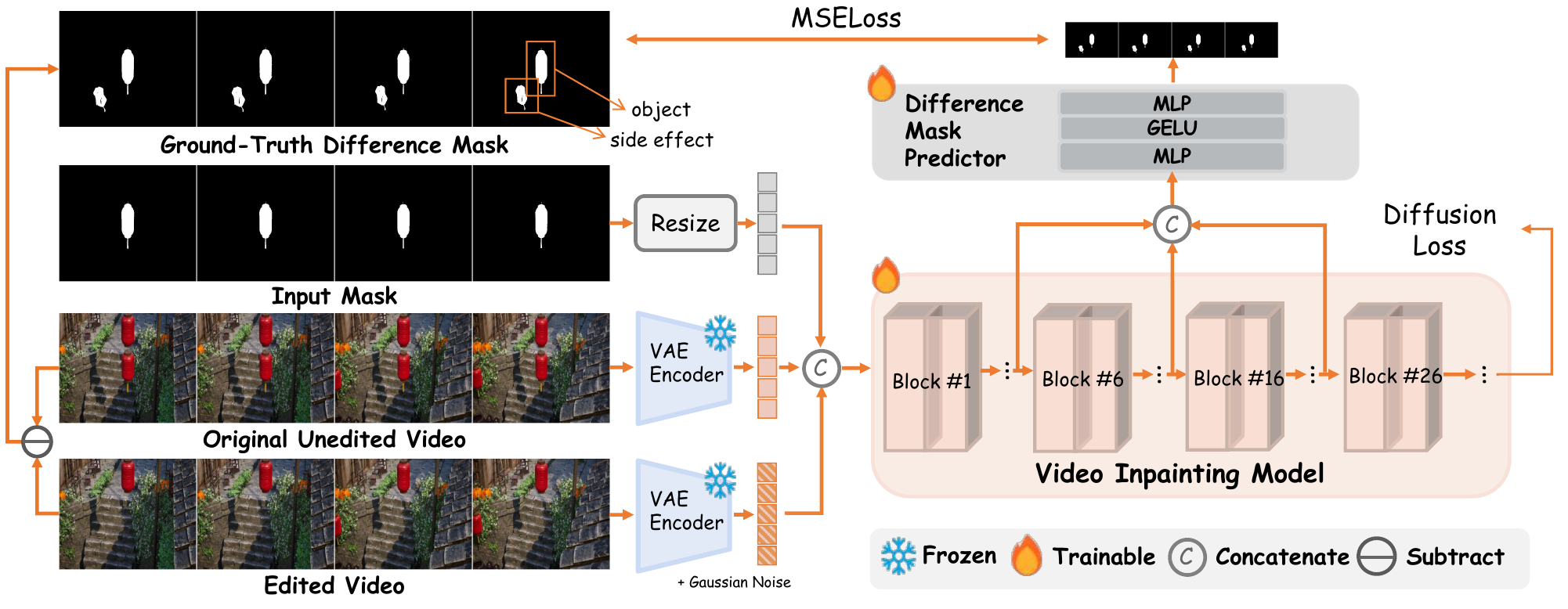}
        \vspace{-8pt}
        \caption{\textbf{The framework of \method}. We concatenate the noisy latents with the original input video and masks, consumed by a video inpainting model. An additional difference mask predictor is introduced to predict the correlated area in video, automatically computed from the input video pairs.}
        \label{training_framework}
        \vspace{-10pt}
    \end{figure}

In this section, we elaborate the model architecture of \method for conducting video object removal, as illustrated in \cref{training_framework}.
In brief, \method is implemented as an inpainting model continued from the foundation video generative models~\cite{wan2025video,kong2025hunyuanvideo} (the Wan2.1 model~\cite{wan2025video} in this paper). 
Following the general architecture in diffusion-based inpainting models~\cite{flux2024,bian2025videopainter}, we extend the model input with the original input video together with object masks.
Distinguished from the typical setting that multiply the mask onto input video, we directly feed the whole video to assist the understanding on environment.
The input masks, with precise boundary generated by 3D engine, are further augmented to enhance model robustness.
To better supervise the model to localize the subtle object-environment interactions, we introduce an additional difference mask predictor to explicitly predict the side effect areas.
We present the detail of \method in the following sub-sections.

\subsection{Reference-based Object Erasing}

We start by formulate the video inpainting task in \method.
Given an input video $\mathcal{V}$ and a binary mask sequence $\mathcal{M}$, where the area of object to be removed is filled with value $1$, the target is to generate an object-erased video $\hat{\mathcal{V}}$. 
For the video condition consumed by the model, most prior methods~\cite{li2025diffueraser,freeformvideoinpainting,zhou2023propainter,videoinpaintingbyjointlylearning} follow a ``mask-and-inpaint'' paradigm, feeding the network with only the non-object area \( \mathcal{V} \odot (1 - \mathcal{M}) \), where $\odot$ indicates point-wise product.
Suppose the noisy latents of diffusion model as $X$, then the model input can be regarded as $[X;\mathcal{V} \odot (1 - \mathcal{M});\mathcal{M}]$.
Such a manner explicitly eliminate the object from input, and is friendly for model convergence.
However, when confronting the side effect removal, isolating the object from the model makes it challenging to localize the object-related region.
In contrast, recent work on image modality has explored to guide the model with the masked region for reference~\cite{jiang2025smarteraser}.
In this paper, we adapt such reference-based erasing, modifying the model input as $[X;\mathcal{V};\mathcal{M}]$.
Experimental results suggest introducing the whole video as guidance significantly increase the performance.
We attribute the advancement that the inner attention mechanism is effective for seeking the inter-region correlations in videos.
Given the object region as input, the model thereby leverage it prior knowledge to localize the side effect regions, thus outperforms the model with masked video input.
Furthermore, the complete video as input serves to enhance the temporal consistency of output video, for introducing the original object-environment interactions.
The visualization comparisons between the two paradigms are shown in \cref{mask_guidance}.
%

\subsection{Mask Augmentation}
\begin{figure}[tbp]
      \centering
      \begin{minipage}{0.48\textwidth}
        \centering
        \includegraphics[width=\linewidth]{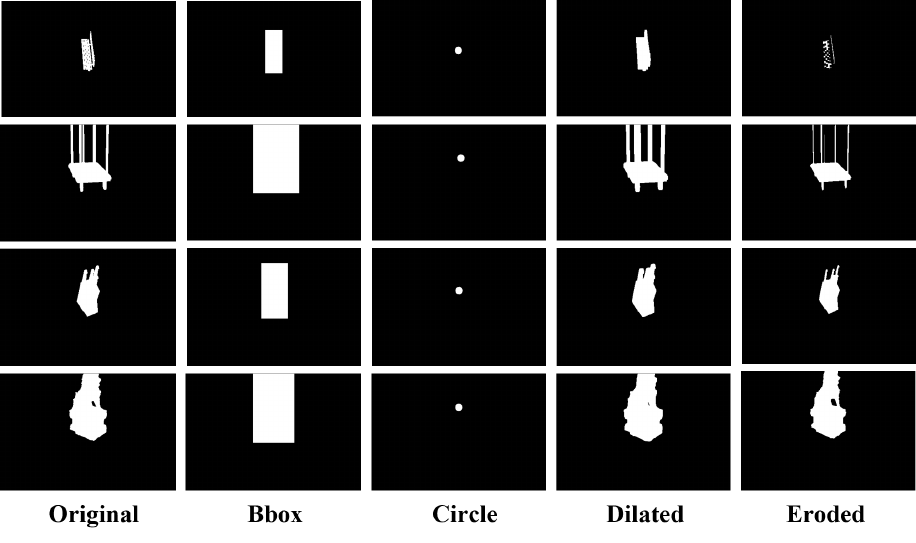}
        \vspace{-5pt}
        \caption{Visualization of various mask augmentation strategies adopted in training.}
        \label{mask_aug}
      \end{minipage}
      \hfill
      \begin{minipage}{0.48\textwidth}
        \centering
        \includegraphics[width=\linewidth]{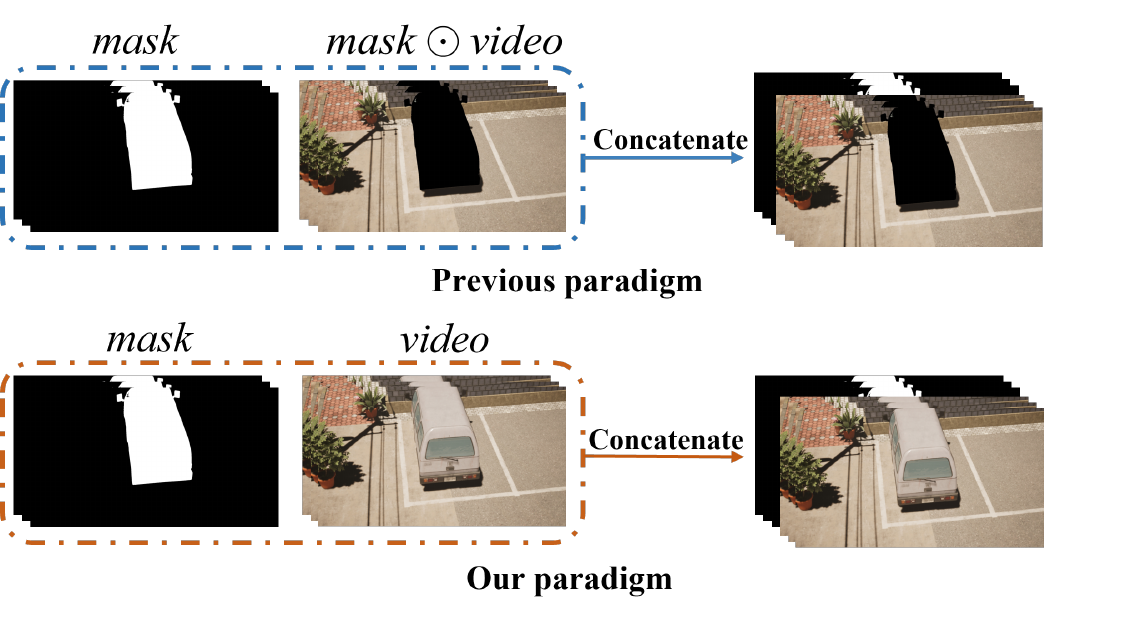}
        \vspace{-8pt}
        \caption{Comparison between the previous paradigm and our reference-based paradigm. 
        }
        \label{mask_guidance}
      \end{minipage}
      \vspace{-11pt}
    \end{figure}

    In real-world applications, user-provided masks often vary in precision, size, and shape—ranging from accurate segmentation maps to coarse bounding boxes or sparse point annotations. 
Since the masks generated by 3D engine is perfectly accurate, training solely on such ideal masks can lead to a performance gap at deployment. To mitigate this, we introduce a set of mask augmentation strategies that simulate diverse mask types likely to appear in practice.
    As shown in \cref{mask_aug}, we adopt five variants: (i) \textit{Original mask}, a precise binary map from ground-truth annotations; (ii) \textit{Point-wise mask}, an extremely sparse point simulating minimal user input; (iii) \textit{Bounding box mask}, a coarse rectangular region enclosing the target; (iv) \textit{Dilated mask}, obtained via morphological dilation to simulate loose annotations; and (v) \textit{Eroded mask}, generated by erosion to mimic under-segmentation.
These variants are randomly sampled during training, which exposes the model to diverse, imperfect masks and improves its generalization to real-world inputs.

\subsection{Explicit Supervision via Difference Mask Prediction}
\label{difference_mask_predictor}
    
Beyond the diffusion loss targeting reconstruct the regions of object and its side effect, we introduce an additional supervision into \method.
Specifically, we inject a \textit{difference mask predictor} into the framework,  predicting binary masks indicating all the areas to be modified in video.
%

The core idea is to leverage the complementary information in the video pairs for training. 
When an object is removed from a scene, it often leaves behind subtle but semantically significant side effects, such as shadows, reflections, and occlusions. 
To explicitly guide the model in attending to these regions, we compute a binary difference mask by comparing the original video \( \mathbf{x}_0 \in \mathbb{R}^{c \times f \times h \times w} \) and its edited counterpart \( \tilde{\mathbf{x}}_0 \). The difference mask \( \mathbf{d}_0 \in \{0,1\}^{f \times h \times w} \) is defined as:

    \vspace{-6pt}
    \begin{equation}
        \mathbf{d}_0^{(t,h,w)} = 
        \begin{cases}
            1, & \text{if } \left\| \mathbf{x}_0^{(t,h,w)} - \tilde{\mathbf{x}}_0^{(t,h,w)} \right\|_2 > \delta \\
            0, & \text{otherwise}
        \end{cases}
    \end{equation}
    
    where \( \delta > 0 \) is a fixed threshold ($\delta=0.09$ in this paper). The resulting binary mask highlights pixel-level differences induced by object removal and is downsampled to match the latent resolution, yielding the ground-truth difference mask \( \mathbf{d}_t \in \{0,1\}^{f \times h/s \times w/s} \).
    
    \paragraph{Difference Mask Predictor.}
    \vspace{-10pt}
    To guide the model in identifying regions influenced by object removal, we design a difference mask predictor $\mathcal{D}_\theta$, which takes as  concatenated token features as input, extracted from multiple transformer blocks. 
Let $\mathbf{x} \in \mathbb{R}^{B \times L \times D_{\text{total}}}$ denote the fused feature sequence, where $L = F_p \times H_p \times W_p$ represents the total number of tokens and $D_{\text{total}}$ is the aggregated channel dimension after selecting and concatenating multiple transformer layers.
The difference mask predictor consists of a two-layer MLP that reduces $D_{\text{total}}$ to a scalar prediction per token. Its output is then reshaped into a 3D spatio-temporal grid with the same shape of video latents:
    \begin{equation}
        \hat{\mathbf{d}}_t = \text{Interpolate}\left(\text{Reshape}(\mathcal{D}_\theta(\mathbf{x})), \text{size} = (F, H, W)\right),
    \end{equation}
    
    where the predicted mask $\hat{\mathbf{d}}_t \in [0,1]^{B \times 1 \times F \times H \times W}$ is upsampled via trilinear interpolation from a coarse patch-level grid $(F_p, H_p, W_p)$ to the full resolution $(F, H, W)$.    
    The module is trained under MSE loss supervision against the ground-truth difference mask $\mathbf{d}_t$ described in Eq.~(3). It functions as an auxiliary self-localization signal to encourage the model to be sensitive to subtle visual effects introduced by object edits.
Then the training objective of \method consists of two terms: the standard diffusion denoising loss and the auxiliary mask prediction loss:
\vspace{-6pt}
\begin{equation}
    \mathcal{L} = \mathbb{E}_{t, \mathbf{z}_0, \boldsymbol{\epsilon}} \left[
        \|\boldsymbol{\epsilon} - \hat{\boldsymbol{\epsilon}}\|_2^2 +
        \lambda \|\hat{\mathbf{d}}_t - \mathbf{d}_t\|_2^2
    \right],
\end{equation}
\vspace{-6pt}

where \( \lambda \) balances the two objectives. This formulation enables the difference mask predictor to guide the model in localizing and identifying regions where object-environment interactions occur.

\section{Experiments}
\label{exp}
\begin{figure}[t]
    \centering
    \includegraphics[width=\linewidth]{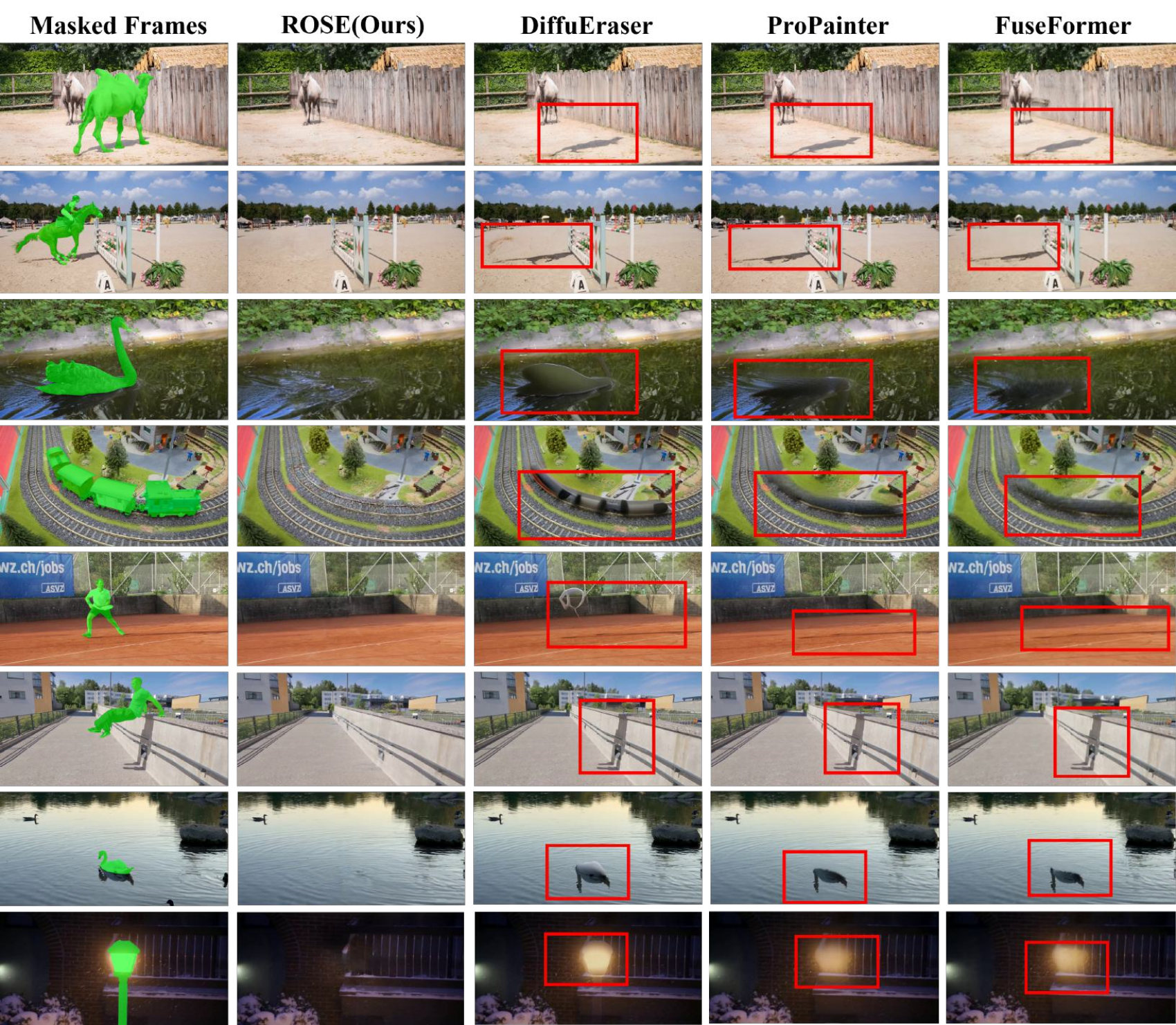}
    \vspace{-5pt}
    \caption{
    Qualitative comparison between our method and existing approaches on real-world samples. Our model demonstrates superior ability and effectively handles complex object-environment interactions, including shadows, reflections, and illumination changes.
    }
    \label{visualization_comparisons}
    \vspace{-17pt}
\end{figure}
\subsection{Experiment Settings}
    \label{experiment_settings}
   \paragraph{Training Data.}
    
Our dataset contains 16,678 synthetic video pairs rendered in Unreal Engine, each 6 seconds (90 frames) at 1920×1080 resolution. It features diverse urban, rural, and natural scenes with dynamic weather, lighting, and interactive objects.

   \paragraph{Evaluation Benchmark and Metrics.}
   \vspace{-10pt}
   \begin{table}[t]
    \vspace{-10pto}
    \centering
    \footnotesize
    \caption{Quantitative comparison on the \textbf{synthetic paired} benchmark (PSNR↑ / SSIM↑ / LPIPS↓)}
    \begin{adjustbox}{max width=\textwidth}
    \begin{tabular}{ll|cccccc}
    \toprule
    \textbf{Category} & \textbf{Metric} & \textbf{ROSE(Ours)} & \textbf{DiffuEraser~\cite{li2025diffueraser}} & \textbf{ProPainter~\cite{zhou2023propainter}} & \textbf{FuseFormer~\cite{liu2021fuseformer}} & \textbf{FloED~\cite{gu2024advanced}} & \textbf{FGT~\cite{zhang2022flow}} \\
    \midrule
    \multirow{3}{*}{Common} 
        & PSNR   & \textbf{36.5998} & 30.9326 & 31.9972 & 31.2325 & 29.8932 & 28.4331 \\
        & SSIM   & \textbf{0.9517} & 0.9204 & 0.9466 & 0.9154 & 0.9066 & 0.8819 \\
        & LPIPS  & \textbf{0.0413} & 0.0825 & 0.0515 & 0.0658 & 0.0738 & 0.0832\\
    \midrule
    \multirow{3}{*}{Shadow}
        & PSNR   & \textbf{33.7876} & 28.9976 & 30.2427 & 28.5520 & 27.8932 & 27.5809 \\
        & SSIM   & 0.9225 & 0.9220 & \textbf{0.9353} & 0.8972 & 0.8834 & 0.8547 \\
        & LPIPS  & 0.0626 & 0.1119 & \textbf{0.0619} & 0.1035 & 0.1208 & 0.1172 \\
    \midrule
    \multirow{3}{*}{Light Source}
        & PSNR   & \textbf{30.0739} & 22.6541 & 23.4291 & 22.8571 & 22.3125 & 21.4579 \\
        & SSIM   & \textbf{0.9209} & 0.8832 & 0.8924 & 0.8630 & 0.8596 & 0.8433 \\
        & LPIPS  & \textbf{0.0862} & 0.1403 & 0.1174 & 0.1410 & 0.1589 & 0.1347 \\
    \midrule
    \multirow{3}{*}{Reflection}
        & PSNR   & \textbf{27.7344} & 26.2914 & 26.9373 & 25.7707 & 25.1018 & 24.3986 \\
        & SSIM   & 0.8715 & 0.8619 & \textbf{0.8763} & 0.8345 & 0.8413 & 0.8421 \\
        & LPIPS  & 0.1129 & 0.1405 & \textbf{0.1072} & 0.1437 & 0.1651 & 0.1520 \\
    \midrule
    \multirow{3}{*}{Mirror}
        & PSNR   & \textbf{28.3498} & 22.1228 & 22.1206 & 22.3175 & 21.3789 & 22.6013 \\
        & SSIM   & \textbf{0.9381} & 0.8855 & 0.8994 & 0.8671 & 0.8678 & 0.8594 \\
        & LPIPS  & \textbf{0.0878} & 0.1751 & 0.1447 & 0.1596 & 0.1845 & 0.1653 \\
    \midrule
    \multirow{3}{*}{Translucent}
        & PSNR   & \textbf{31.4264} & 28.4520 & 29.8910 & 28.1712 & 27.3924 & 27.4802 \\
        & SSIM   & \textbf{0.9470} & 0.9259 & 0.9397 & 0.9168 & 0.8956 & 0.9034 \\
        & LPIPS  & \textbf{0.0598} & 0.1036 & 0.0722 & 0.0914 & 0.1134 & 0.1210 \\
    \midrule
    \midrule
    \multirow{3}{*}{\textbf{Mean}}
        & PSNR   & \textbf{31.1221} & 26.5024 & 27.1991 & 26.2566 & 25.4847 & 25.2353 \\
        & SSIM   & \textbf{0.9170} & 0.8981 & 0.9148 & 0.8795 & 0.8697 & 0.8641 \\
        & LPIPS  & \textbf{0.0772} & 0.1284 & 0.0946 & 0.1208 & 0.1324 & 0.1289 \\
    \bottomrule
    \end{tabular}
    \end{adjustbox}
    \label{exp_rose}
    \end{table}
    \vspace{-10pt}
   Existing benchmarks in the video inpainting domain mainly suffer two limitations. First, most of them lack access to paired edited videos following real-world physical rules, which restricts quantitative evaluation due to the absence of ground-truth. Second, they overlook the \emph{side effects} induced by object-environment interactions hat are critical for assessing the semantic correctness and realism of inpainting. Consequently, these benchmarks fail to capture fine-grained challenges that frequently arise in real-world applications.
\begin{wraptable}[21]{r}{0.5\textwidth}
\vspace{-10pt}
\centering
\scriptsize
\caption{Ablation study on ROSE-Bench (PSNR ↑ / SSIM ↑ / LPIPS ↓)}
\label{ablation_wrap}
\begin{adjustbox}{max width=0.48\textwidth}
\begin{tabular}{ll|cccc}
\toprule
\textbf{Category} & \textbf{Metric} & Base & +MRG & +MA & +DMP \\
\midrule
\multirow{3}{*}{Common} 
    & PSNR   & 32.58 & 35.24 & 35.37 & 36.60 \\
    & SSIM   & 0.937 & 0.950 & 0.948 & 0.952 \\
    & LPIPS  & 0.053 & 0.040 & 0.041 & 0.041 \\
\multirow{3}{*}{Shadow}
    & PSNR   & 30.65 & 33.29 & 33.62 & 33.79 \\
    & SSIM   & 0.914 & 0.920 & 0.921 & 0.923 \\
    & LPIPS  & 0.081 & 0.061 & 0.061 & 0.063 \\
\multirow{3}{*}{Light}
    & PSNR   & 24.99 & 30.37 & 25.80 & 30.07 \\
    & SSIM   & 0.894 & 0.923 & 0.893 & 0.921 \\
    & LPIPS  & 0.112 & 0.074 & 0.107 & 0.086 \\
\multirow{3}{*}{Reflect.}
    & PSNR   & 25.39 & 27.71 & 27.08 & 27.73 \\
    & SSIM   & 0.836 & 0.843 & 0.841 & 0.872 \\
    & LPIPS  & 0.131 & 0.109 & 0.122 & 0.113 \\
\multirow{3}{*}{Mirror}
    & PSNR   & 22.63 & 28.45 & 27.37 & 28.35 \\
    & SSIM   & 0.905 & 0.941 & 0.921 & 0.938 \\
    & LPIPS  & 0.142 & 0.076 & 0.107 & 0.088 \\
\multirow{3}{*}{Translucent}
    & PSNR   & 27.43 & 30.98 & 30.12 & 31.43 \\
    & SSIM   & 0.925 & 0.949 & 0.947 & 0.947 \\
    & LPIPS  & 0.087 & 0.052 & 0.056 & 0.060 \\
\midrule
\multirow{3}{*}{\textbf{Mean}}
    & PSNR   & 27.28 & 30.84 & 29.89 & 31.12 \\
    & SSIM   & 0.902 & 0.918 & 0.912 & 0.917 \\
    & LPIPS  & 0.101 & 0.071 & 0.082 & 0.077 \\
\bottomrule
\end{tabular}
\end{adjustbox}
\label{ablation}
\vspace{-10pt}
\end{wraptable}

To address these gaps, we construct \textit{ROSE-Bench}, a comprehensive evaluation benchmark on video object removal, consisting of following subsets: 

(i) \textit{Synthetic paired} benchmark tailored for evaluation under diverse physical interaction effects. Using the same simulation approach described in \cref{dataset_construct}, the benchmark consists of 6 representative categories: common, light source, mirror, reflection, shadow, and translucent, each modeling a specific class of object-environment interaction. Every category contains 10 high-quality triplets of video sequences, \textit{i.e.,} original, edited, and mask videos, offering precise and controllable evaluation of model behavior under different side-effect conditions.

    


 (ii) \textit{Realistic paired} benchmark constructed using a \textit{copy-and-paste} strategy based on the video segmentation dataset dataset DAVIS~\cite{davis}. We copy a masked object from one video into another. The resulting video with inserted object is treated as input, while the original unaltered video serves as the ground-truth. 
This process allows us to construct realistic and diverse test cases that mirror practical editing scenarios while preserving access to ground-truth supervision. 
For quantitative evaluation on paired benchmark, we compute PSNR~\cite{psnr}, SSIM~\cite{ssim}, and LPIPS~\cite{lpips} across both synthetic and real-world test sets. These metrics capture both low-level structural fidelity and perceptual similarity, assessing the model performance under various side-effect challenges.

(iii) \textit{Realistic unpaired} benchmark containing real videos with masks. Different from the second subset, we directly feed real-world videos into model, which are also sampled from DAVIS~\cite{davis}. To conduct evaluation without ground-truth, we select related metrics from the VBench~\cite{huang2023vbench}, a widely-adopted benchmark on text-to-video generation, for evaluating the quality of  output videos on motion smoothness, background consistency and temporal flickering.

\vspace{-10pt}
    
\paragraph{Implementation Details.}

In the training process, we resize all the video pairs into the resolution of $720\times 480$ and use $81$ frames for training. 
The backbone model is a controllable generation variant of Wan2.1 1.3B version~\cite{wan2025video}. We fully train the model together with the difference mask predictor in $80000$ optimization steps with $0.00002$ learning rate on $4$ NVIDIA H800 GPUs.

\begin{wraptable}[13]{r}{0.5\textwidth}
    \centering
    \scriptsize 
    \caption{Quantitative comparison on \textbf{realistic paired} benchmark.}
    \label{tab:realworld_metrics}
    \begin{tabular}{lccc}
    \toprule
    \textbf{Method} & \textbf{PSNR} ↑ & \textbf{SSIM} ↑ & \textbf{LPIPS} ↓ \\
    \midrule
    \textbf{ROSE(Ours)}         & 31.34 & \textbf{0.923} & \textbf{0.092} \\
    DiffuEraser~\cite{li2025diffueraser} & 29.97 & 0.901 & 0.128 \\
    ProPainter~\cite{zhou2023propainter} & \textbf{32.81} & 0.917 & 0.122 \\
    FuseFormer~\cite{liu2021fuseformer}  & 26.52 & 0.885 & 0.151 \\
    FloED~\cite{gu2024advanced}          & 28.48 & 0.881 & 0.147 \\
    FGT~\cite{zhang2022flow}            & 27.53 & 0.874 & 0.135 \\
    \bottomrule
    \end{tabular}
    \label{exp_copy_paste}
\end{wraptable}

\subsection{Comparisons with Previous Methods}
    \vspace{-5pt}
    \paragraph{Quantitative Evaluation.}
    For quantitative evaluation, we compare our method with flow-based transformers (ProPainter~\cite{zhou2023propainter}, FuseFormer~\cite{liu2021fuseformer}, FGT~\cite{zhang2022flow}) and diffusion-based methods (DiffuEraser~\cite{li2025diffueraser}, FLoED~\cite{gu2024advanced}). We evaluate all methods on the three components of ROSE-Bench: synthetic paired benchmark (\cref{exp_rose}), realistic paired benchmark (\cref{exp_copy_paste}), and real-world videos (\cref{vbench_eval}). Our model achieves superior performance in object removal, as measured by PSNR~\cite{psnr}, SSIM~\cite{ssim}, and LPIPS~\cite{lpips}, and excels in maintaining motion smoothness, background consistency, and subject consistency in \cref{vbench_eval}.

    \begin{table}[htbp]
    \centering
    \caption{
    VBench-based evaluation on the \textbf{realistic unpaired} benchmark. (Best scores are \textbf{bolded}).
    }
    \label{tab:vbench_full}
    \small
    \begin{adjustbox}{max width=\textwidth}
    \begin{tabular}{lccccc}
    \toprule
    \multirow{2}{*}{Method} & 
    \multirow{2}{*}{\begin{tabular}[c]{@{}c@{}}\textbf{Motion}\\ \textbf{Smoothness ↑}\end{tabular}} &
    \multirow{2}{*}{\begin{tabular}[c]{@{}c@{}}\textbf{Background}\\ \textbf{Consistency ↑}\end{tabular}} &
    \multirow{2}{*}{\begin{tabular}[c]{@{}c@{}}\textbf{Temporal}\\ \textbf{Flickering ↓}\end{tabular}} &
    \multirow{2}{*}{\begin{tabular}[c]{@{}c@{}}\textbf{Subject}\\ \textbf{Consistency ↑}\end{tabular}} &
    \multirow{2}{*}{\begin{tabular}[c]{@{}c@{}}\textbf{Imaging}\\ \textbf{Quality ↑}\end{tabular}} \\
    \\
    \midrule
    DiffuEraser               & 0.972 & 0.902 & \textbf{0.931} & 0.891 & \textbf{0.658} \\
    ProPainter                & \textbf{0.975} & 0.917 & 0.932 & 0.903 & 0.626 \\
    FuseFormer                & 0.971 & 0.905 & 0.938 & 0.892 & 0.625  \\
    FloED                    &  0.973 & 0.904 & 0.932 & 0.889 & 0.618  \\
    FGT                      &  0.971 & 0.897 & 0.933 & 0.895 & 0.614  \\   
   \rowcolor[gray]{0.95} \method \textbf{(Ours)}      & \textbf{0.975} & \textbf{0.923} & 0.936 & \textbf{0.908} & 0.630 \\
    \bottomrule
    \end{tabular}
    \end{adjustbox}
    \label{vbench_eval}
\end{table}
    
    \vspace{-10pt}
    \paragraph{Qualitative Evaluation.} 
    
    For qualitatitve evaluation, we compare our method with ProPainter~\cite{zhou2023propainter}, FuseFormer~\cite{liu2021fuseformer} and DiffuEraser~\cite{li2025diffueraser}. Qualitative visualization results can be seen in \cref{visualization_comparisons}. In the \cref{visualization_comparisons}, we demonstrate cases with various different side effects like shadows, reflection and lumination changes and we can obviously find that our model shows superior performance over other methods. The side effects areas that previous works fail to fill in have been framed in red boxes.

\subsection{Ablation Study}
   
    We perform ablation studies to demonstrate the effectiveness of our designs. We keep training settings same as in \cref{experiment_settings} to ensure the fairness of comparisons and we evaluate our methods on the synthetic paired benchmark. 
    We set the baseline with the following settings: use the "mask-and-inpaint" paradigm, without mask augmentation and difference mask predictor. And in \cref{ablation}, MRG stands for mask region guidance, MA stands for mask augmentation and DMP stands for difference mask predictor. In \cref{ablation}, we have shown that our primary designs are effective and useful.

\section{Discussion}
\label{discussion}
    This paper introduces \method, a unified framework for video object removal that addresses both target objects and their side effects, such as shadows, reflections, and lighting distortions. By leveraging synthetic data from a 3D rendering pipeline, we alleviate real-world data scarcity while ensuring diverse scenes and camera motions. Our diffusion transformer architecture excels in object localization and side effect removal via differential mask supervision. The proposed ROSE-Bench offers systematic evaluation for object-environment interactions, addressing a key gap in video inpainting. Extensive experiments show that \method significantly outperforms prior methods and generalizes well to real-world videos. These contributions advance video editing and set new benchmarks for handling complex visual artifacts. Future work will explore real-time optimization and broader environmental effects to further bridge synthetic and real-world domains. Despite its strengths, \method has \textbf{limitations}:
(1) It may produce flickering artifacts under large motion, as shown in \cref{vbench_eval};
(2) Inference time grows with video length, reducing efficiency on long sequences.

\medskip

\newpage
\clearpage
\small{\bibliographystyle{plain}
\bibliography{references}}


\clearpage
\newpage

\end{document}